\DocumentMetadata{
	lang		= eng-US, 	% default
	pdfstandard	= A-3u,		% A-2b, A-2u, A-3b, A-3u. Check that your figures also meet the standard you choose.
	pdfversion	= 1.7, 		% an older version of PDF, but very widely supported
}

\documentclass[colorlinks,upint,subscriptcorrection,varvw,hyphenate,balance,greek,russian,vietnamese,german,nofoot]{asmeconf} % pdftex

\usepackage{algorithm}
\usepackage{algpseudocode}

\hypersetup{%
	pdfauthor={John H. Lienhard},									  % <=== change to YOUR name
	pdftitle={ASME Conference Paper LaTeX Template},                  % <=== change to YOUR pdf file title
	pdfkeywords={ASME conference paper, LaTeX template, BibTeX style},% <=== change to YOUR pdf keywords
	pdfsubject = {Describes the asmeconf LaTeX template},			  % <=== change to YOUR subject
}
\allowdisplaybreaks % from amsmath package, allows multiline equations to break across pages (delete if not wanted)
\begin{document}

% Change these fields to the right content for your conference.
% You can comment these out if for some reason you don't want a header.
% Use title case for the conference name (first letters capitalized), not all capitals

\ConfName{}
\ConfAcronym{}
\ConfDate{} 
\ConfCity{}
\PaperNo{}

% Units of measure (e.g., cm) and other specialty lowercase terms in the title should be 
%   enclosed in \NoCaseChange{...} to maintain lower case type.
%   The rest of the title will automatically be set in all capital letters.
%
%	\title{Place Title Here: Place Subtitle After Colon} 

\title{Generative Multi-Robot Motion Planning via Diffusion Modeling with Multi-Agent Reinforcement Learning Guidance
} % <=== replace with YOUR title
 
%   Put author names into the order you want. Use the same order for affiliations.
%   \affil{#} tags the author's affiliation to the address in \SetAffiliation{#}.
%   No space between last name and \affil{#}, separate names with commas.
%
%	For a sole author or a single affiliation for all authors, {#} may be left empty, i.e. \affil{} and \SetAffiliation{} (but not with [grid] option!)
%
%   \CorrespondingAuthor{email} follows that author's affiliation, no spaces before or after 
%   If multiple corresponding authors, put both email addresses in the same command and place after both authors.
%
%   \JointFirstAuthor, if applicable, follows the affiliation of the relevant authors, no spaces.

\SetAuthors{%
    Suk Ki Lee\affil{1}\JointFirstAuthor, 
	Venkata Sai Deepak Mutta\affil{1}\affil{2}\JointFirstAuthor, 
	Hyunwoong Ko\affil{1}\CorrespondingAuthor{Hyunwoong.Ko@asu.edu}
	}
%	Note: Luis and Maria are not real people.  Henry and Catherine have been dead for >450 years.

\SetAffiliation{1}{School of Manufacturing Systems and Networks, Arizona State University, Mesa, AZ}
\SetAffiliation{2}{Michael W. Hall School of Mechanical Engineering, Mississippi State University, Starkville, MS}

%   Note: You can force a line break in the address using \\ 

%	To switch from inline author names to gridded names, use the [grid] option.

\maketitle

%%% Use this footnote for tracking various versions of your draft. Change text to suit your own needs. 
%%% \date{..} calls the same command. 
%\versionfootnote{Documentation for \texttt{asmeconf.cls}: Version~\versionno, \today.}% <=== Delete before final submission.

%%% Change the following to your keywords.  Keywords are automatically printed at the end of the abstract.
%%% This command MUST COME BEFORE the end of the abstract.
%%% If you don't want keywords, leave the argument of \keywords{} empty (or use the abstract* environment)

\keywords{Multi-Robot, Motion Planning, Multi-Agent Reinforcement Learning, Diffusion Model, Guidance Method}

%%%%%  End of fields to be completed. Now write your paper. %%%%%%%%%%%%%%%%%%%%%%%%%%%%%%%%%%%%%%%%%%%

%%%%%  ABSTRACT  %%%%%%%%%%%%%%%%%%%%%%%%%%%%%%%%%%%%%%%%%%%%%%%%%%%
%%
%% Abstract should be 200 words or less
\begin{abstract}
Coordinating multiple robots in shared environments requires generating feasible trajectories for each agent while accounting for interactions among agents.
Centralized planning approaches become difficult to scale as the number of robots increases, while decentralized approaches that allow each agent to plan independently do not inherently account for inter-agent interactions.
This paper presents a framework for coordinated multi-robot motion planning that combines decentralized generative trajectory planning with multi-agent reinforcement learning (MARL)-based coordination.
Each robot independently generates candidate trajectories using a diffusion model trained on single-agent motion data, leveraging the generative model's ability to produce feasible and diverse trajectories.
To reduce conflicts between agents, a centralized value function trained via MARL guides the reverse diffusion process through gradient-based steering, enabling interaction-aware trajectory generation without centralized joint planning or retraining of the generative model.
This guidance follows an exponential tilting formulation, in which the value function biases the denoising distribution toward trajectories with higher expected multi-agent return.
The framework is evaluated in a simulated maze environment with four mobile robots.
Experimental results show that the proposed value-guided diffusion planning reduces the inter-agent interference rate from 55.4\% to 41.8\%, demonstrating that coordination can be effectively achieved while preserving the scalability of decentralized trajectory generation.
These results suggest that MARL-based value guidance can effectively introduce coordination into decentralized generative planners without requiring a fully joint multi-robot model.
\end{abstract}

%%%%%%%%%  NOMENCLATURE (OPTIONAL) %%%%%%%%%%%%%%%%%%%%%%%%%%%%%%%%%
%%
%% To change space between the symbols and  definitions, use \begin{nomenclature}[Xcm] where X is a number 
%% The unit cm can be replaced by any LaTeX unit of dimension: pt, in, ex, em, pc, etc.
%% Default is 2em.
%%
%% \EntryHeading{..} produces an italicized subheading in the nomenclature list, e.g., \EntryHeading{Greek letters}

%\begin{nomenclature}
%\EntryHeading{Roman letters}
%\entry{$k$}{Thermal conductivity [W m$^{-1}$ K$^{-1}$]}
%\entry{$\vec{q}$}{Heat flux vector [W m$^{-2}$]}
%\end{nomenclature}

%%%%%%%%%  BODY OF PAPER %%%%%%%%%%%%%%%%%%%%%%%%%%%%%%%%%

\section{Introduction}
%Introducing Multi-robot Systems
Robotic systems are increasingly deployed in environments that require complex and coordinated tasks. 
As a result, conventional single-robot approaches are giving way to multi-robot configurations in which multiple agents operate simultaneously within a shared workspace. 
In many industrial settings, especially manufacturing, robots must coordinate their motions while pursuing individual task objectives, enabling parallel task execution and flexible system operation \cite{arai2002advances, saha2006multi, lee2025generative}.
As manufacturing systems grow in complexity and production demands increase, effective multi-agent coordination becomes
essential. 
Without proper coordination, robot trajectories may conflict or compete for shared space, leading to delays,
inefficient motion, or physical interference. 
Ensuring that multiple robots can move efficiently while avoiding such conflicts, therefore, remains a fundamental challenge in multi-robot motion planning \cite{arai2002advances, lee2025generative, stone2025safezone,
poudel2020heuristic}.

%Motion Planning - Centralized, Decentralized
Coordinating multiple robots in a shared environment is challenging because each robot must plan its motion not only to accomplish its own task but also to account for the trajectories of other agents.
Centralized planning approaches, which jointly optimize trajectories for all agents, can explicitly model such
interactions \cite{arai2002advances, sharon2015conflict}.
However, as the number of robots and task complexity increase, the joint planning space grows rapidly, leading to significant computational complexity and limited scalability \cite{poudel2023decentralized}.
Decentralized approaches instead allow each agent to plan independently based on local information, offering greater scalability and resilience to uncertainty \cite{poudel2023decentralized, chen2017decentralized}.
However, independently generated trajectories do not inherently account for interactions among agents, and conflicts may still arise when robots operate in close proximity within a shared workspace.
Addressing this tension requires a planning framework that preserves the independence of decentralized generation while still enabling coordination among agents.

% Generative ML, Diffusion
Recent advances in generative machine learning offer a promising direction for addressing these challenges.
Unlike deterministic planners that produce a single trajectory, generative models learn a distribution over feasible motions from data, enabling diverse and uncertainty-aware trajectory generation \cite{lee2025generative,lee2025generative2,ho2020denoising,janner2022planning}.
In particular, diffusion models have demonstrated strong performance in trajectory generation tasks by capturing
complex motion distributions through an iterative denoising process \cite{janner2022planning, carvalho2023motion}.
This generative perspective allows each agent to independently sample feasible trajectories conditioned on its own task constraints, without requiring a centralized planner. 
However, because agents plan independently using the same generative model, their trajectories may still conflict
in shared space.
This motivates the need for a coordination mechanism that can guide trajectory generation while preserving the
scalability of decentralized planning, and without requiring retraining of the generative model.

% Our Framework (diffusion + MARL)
To address this need, this paper proposes a framework for coordinated multi-robot motion planning that combines
decentralized generative trajectory planning with reinforcement learning-based coordination.
Each robot independently generates candidate trajectories using a diffusion-based generative model, preserving the scalability advantages of decentralized planning \cite{janner2022planning, carvalho2023motion}.
To account for interactions among agents in shared spaces, coordination signals derived from multi-agent reinforcement learning are incorporated during the trajectory generation process \cite{zhang2021multi, lowe2017multi, schulman2017proximal, yu2022surprising}.
This approach enables decentralized trajectory generation while guiding agents toward interaction-aware behaviors,
reducing inter-agent interference without requiring centralized joint planning or retraining of the generative model \cite{janner2022planning}.

% Remainder
The remainder of this paper is organized as follows.
Section~\ref{sec:litreview} first reviews related work on diffusion models for trajectory generation, reinforcement
learning for multi-agent coordination, and the broader challenges in multi-robot motion planning.
Section~\ref{sec:problem} then formalizes the problem of coordinated multi-robot motion planning considered in
this work.
Building on this formulation, Section~\ref{sec:method} presents the proposed framework, including single-agent
diffusion-based planning, its extension to decentralized multi-agent settings, and the incorporation of a
MARL-based centralized value function for coordination.
Section~\ref{sec:experiments} next describes the experimental setup, evaluation metrics, and scenarios used to assess the framework.
The results are presented in Section~\ref{sec:results}, where both quantitative and qualitative analyses are
provided and discussed.
Finally, Section~\ref{sec:conclusion} summarizes the main contributions of this work and outlines directions for
future research.

%%%%%%%%%%%%%%%%%%%%%%%%%%%%%%%%%%%%%%%%%%%%%%%%%%%%%%%%%%%

\section{Literature Review}
\label{sec:litreview}
This section reviews prior work on multi-robot motion planning, diffusion-based trajectory generation, and reinforcement learning for multi-agent coordination, establishing the context and motivation for the proposed
framework.

\subsection{Multi-Robot Motion Planning}
\label{sec:litreview_multirobot}

Multi-robot motion planning addresses the problem of generating coordinated trajectories for multiple agents operating simultaneously in a shared environment \cite{banik2025path,poudel2023decentralized}.
A fundamental distinction in existing approaches is between centralized and decentralized planning strategies.
Centralized methods jointly optimize trajectories for all agents, enabling explicit modeling of inter-agent interactions. Classical formulations of this problem have been extensively studied in the multi-agent path finding (MAPF) literature
\cite{silver2005cooperative, ma2017multi, sharon2015conflict}.
However, such approaches become increasingly difficult to scale as the number of robots grows \cite{poudel2023decentralized}.

Decentralized methods address this scalability challenge by allowing each agent to plan independently, but do not inherently account for interactions among agents, leaving open the problem of ensuring compatibility among independently planned trajectories \cite{poudel2023decentralized}.
These challenges are particularly prominent in multi-robot manufacturing environments, where robots must coordinate their motions in shared workspaces while satisfying task-level constraints such as sequencing, reachability, and process dependencies \cite{arai2002advances, poudel2023decentralized}.
Prior work in cooperative additive manufacturing has shown that while centralized planners can produce optimal schedules for small-scale problems, their performance degrades as task complexity and the number of robots increase, motivating the need for more scalable coordination strategies \cite{poudel2023decentralized, stone2025safezone}.
These observations motivate the development of planning frameworks that preserve the scalability of decentralized planning while incorporating effective mechanisms for inter-agent coordination.

\subsection{Diffusion-Based Trajectory Generation}
\label{sec:litreview_diffusion}
Recent advances in generative modeling have introduced diffusion models as a powerful framework for learning
complex data distributions.
Diffusion models generate samples through an iterative denoising process that gradually transforms noise into
structured data, enabling the modeling of high-dimensional distributions \cite{ho2020denoising, song2020denoising}.
Unlike deterministic planners that produce a single solution, diffusion models capture a distribution over feasible outputs, enabling diverse and uncertainty-aware generation \cite{ho2020denoising, song2020denoising, lee2025generative}.
Building on these developments, recent work has explored the use of diffusion models for trajectory generation in
robotics and control \cite{lee2025generative}.

Janner et al. proposed Diffuser, a framework that formulates trajectory planning as a diffusion process over the full trajectory space \cite{janner2022planning}.
By learning a distribution over expert trajectories, Diffuser enables goal-conditioned sampling through an inpainting mechanism that fixes the start and goal states during the reverse diffusion process.
Importantly, Diffuser also demonstrated that inference-time value guidance can be used to steer sampled trajectories toward higher-reward regions, providing a principled mechanism for incorporating task objectives without retraining the generative model \cite{janner2022planning, dhariwal2021diffusion}.
This value-guided sampling approach draws on classifier guidance techniques from image generation, where gradients of a learned value function are used to bias the denoising process \cite{dhariwal2021diffusion}.

Extending this line of work to robotics settings, subsequent work has extended diffusion-based planning to robotics settings.
Carvalho et al. proposed Motion Planning Diffusion, which learns trajectory priors from robot motion data and samples from the posterior distribution conditioned on task goals, demonstrating strong generalization to environments not seen during training \cite{carvalho2023motion}.
Chi et al. introduced Diffusion Policy, representing visuomotor robot policies as conditional diffusion processes and demonstrating the ability to capture multimodal action distributions across a range of manipulation tasks \cite{chi2025diffusion}.

While these works demonstrate the effectiveness of diffusion models for single-agent trajectory generation and policy learning, their direct extension to multi-agent settings remains limited. 
Generating coordinated trajectories for multiple agents using diffusion models requires mechanisms that account for inter-agent interactions during the generation process, which is not addressed by existing single-agent formulations.

\subsection{Reinforcement Learning for Multi-Agent Coordination}
\label{sec:litreview_marl}

Reinforcement learning (RL) provides a framework for learning agent behavior through interaction with an environment, enabling the development of policies that optimize long-term objectives without requiring explicit models of system dynamics \cite{sutton1998reinforcement}.
In single-agent settings, proximal policy optimization (PPO) has demonstrated strong performance across a wide
range of continuous control and robot navigation tasks \cite{schulman2017proximal}.

Extending RL to multi-agent settings introduces additional challenges, including non-stationarity of the environment and the need for coordination among agents \cite{zhang2021multi, busoniu2008comprehensive}.
Multi-agent reinforcement learning (MARL) methods address these challenges by learning policies that account for the behaviors of other agents. 
A widely adopted paradigm is centralized training with decentralized execution, in which a centralized critic observes the global state during training to provide coordinated value estimates, while each agent executes its policy independently at deployment time \cite{lowe2017multi}.
This centralized value function encodes inter-agent interaction patterns learned through experience, providing a coordination signal that reflects the collective behavior of the group. 
Yu et al. demonstrated that MAPPO, a multi-agent extension of PPO under the centralized training with decentralized execution paradigm, achieves strong performance in cooperative multi-agent tasks, establishing it as an effective approach for learning centralized value functions in multi-robot settings \cite{yu2022surprising}.

While MARL-based value functions have shown promise for learning coordination policies, their integration with generative trajectory models for multi-robot motion planning remains underexplored. 
The proposed framework addresses this gap by incorporating a centralized MARL value function as a guidance signal during the diffusion-based trajectory generation process, enabling coordination without centralized joint planning or retraining of the generative model.

%%%%%%%%%%%%%%%%%%%%%%%%%%%%%%%%%%%%%%%%%%%%%%%%%%%%%%%%%%%
\section{Problem Statement}
\label{sec:problem}
% P1: How can trajectory generation be decentralized while maintaining feasible and compatible motions for multiple robots?
Coordinating multiple robots operating in a shared environment remains a fundamental challenge in multi-agent motion planning.
When multiple robots navigate simultaneously, their trajectories may interfere with each other, leading to delays, conflicts, or inefficient motion.
While centralized planning can explicitly account for interactions between agents, the complexity of coordination increases as the number of robots grows.
A scalable alternative is to allow each agent to independently generate its own motion plan, avoiding the need to reason over a combined agent state.
However, enabling decentralized trajectory generation while still producing feasible and diverse motions that remain compatible with the trajectories of other robots operating in the same environment and satisfying their own task constraints remains a key challenge.

% P2: How can independently generated trajectories be coordinated to reduce inter-agent interference in shared environments?
Even when trajectories are generated independently and remain feasible for individual robots, interactions between robots may still produce conflicts during execution.
Independently generated trajectories may intersect or compete for shared workspace, leading to inter-agent interference that degrades overall system performance.
Coordinating such independently generated trajectories to reduce conflicts in shared environments, therefore
becomes another central challenge.

%%%%%%%%%%%%%%%%%%%%%%%%%%%%%%%%%%%%%%%%%%%%%%%%%%%%%%%%%%%

%%%%%%%%%%%%%%%%%%%%%%%%%%%%%%%%%%%%%%%%%%%%%%%%%%%%%%%%%%%
\section{Methodology}
\label{sec:method}

\begin{figure*}[thbp]
    \centering
    \includegraphics[width=\textwidth]{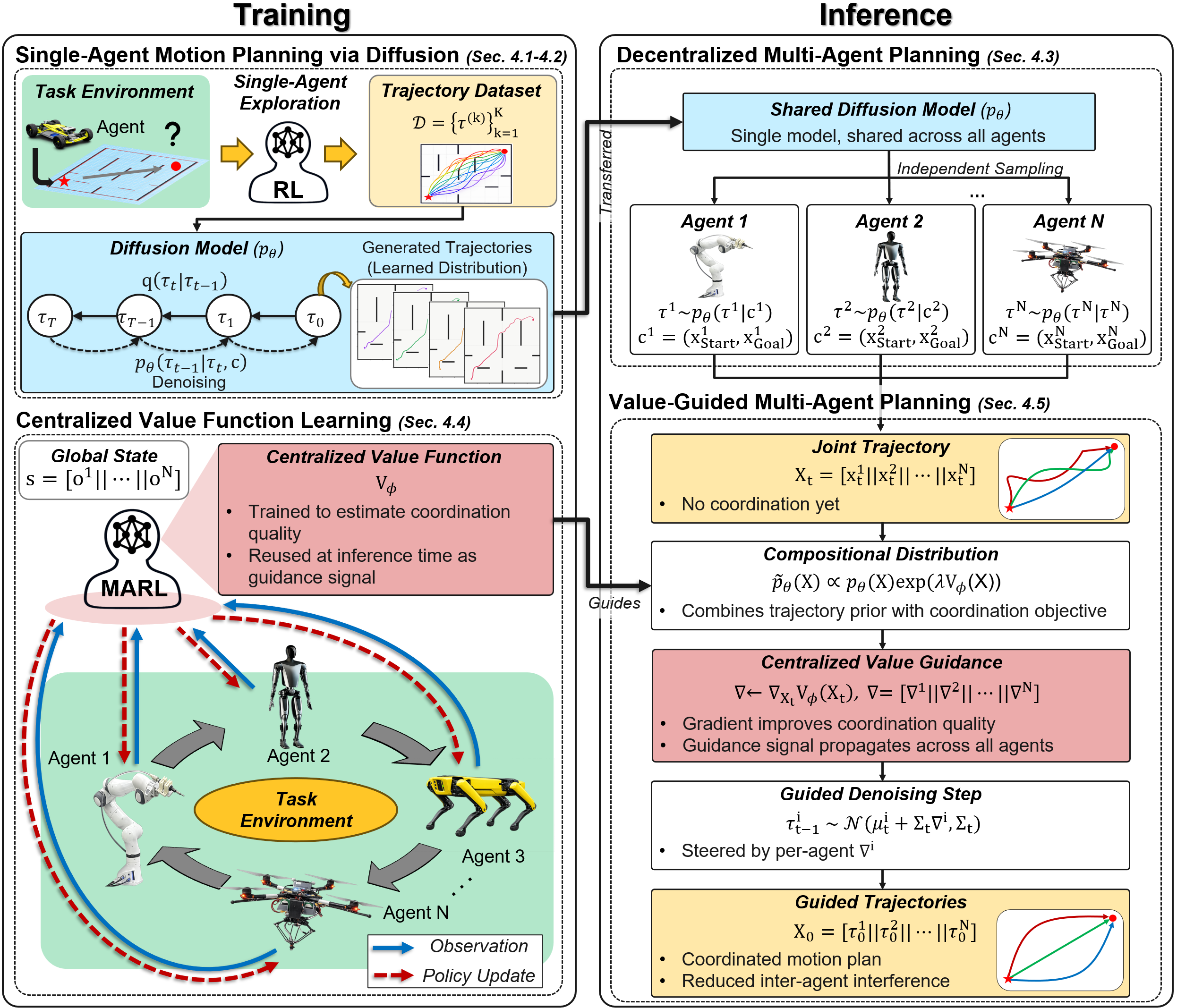}
    \caption{Overall architecture of the proposed framework. In the single-agent data collection phase, an RL policy learns to solve the task in the environment, and the resulting diverse trajectories are used to train a diffusion model that learns the underlying motion distribution, enabling goal-conditioned trajectory generation for a single agent. In the centralized value function learning phase, a value function is trained via multi-agent RL (MARL) to estimate coordination quality from the global state of all agents, with learned feedback propagated back to each agent's policy. At inference time, the trained diffusion model is shared across all agents without retraining, and each agent independently generates a trajectory conditioned on its own start-goal pair. The independently generated trajectories are then guided by the value function gradient at each denoising step, steering the sampling process toward coordinated multi-agent behaviors and producing the final trajectory set with reduced inter-agent interference.
    }
    \label{fig:framework}
\end{figure*}

This section presents the proposed framework for generative multi-agent motion planning via diffusion modeling with multi-agent RL guidance.
At a high level, our approach first generates candidate trajectories using a diffusion model and then refines them using a centralized value function that promotes coordinated multi-agent behaviors.
The overall framework consists of four components:
(1) single-agent motion planning via diffusion, where a diffusion-based motion planner generates feasible trajectories for a single agent under start-goal constraints,
(2) decentralized multi-agent planning, where multiple agents independently sample trajectories using the same diffusion model,
(3) a centralized value function for coordination that evaluates the coordination quality of joint multi-agent trajectories, and
(4) value-guided multi-agent planning, where the diffusion sampling procedure is guided by the value function to steer trajectory generation toward more cooperative behaviors.
An overview of the proposed framework is illustrated in Fig.~\ref{fig:framework}.
The following subsections describe each component in detail.
The proposed framework constructs and refines motion trajectory distributions through a sequence of stages: (i) trajectory distribution construction via reinforcement learning, (ii) generative modeling using diffusion, (iii) decentralized multi-agent trajectory generation, and (iv) coordination through value-guided sampling.
%%%%%%%%%%%%%%%%%%
 
%%%%%%%%%%%%%%%%%%

%%%%%%%%%%%%%%%%%%%%%%%%%%%%%%%%%%%%%%%%%%%%%%%%%%%%%%%%%%%
\subsection{Single-Agent Policy Learning for Trajectory Distribution Construction}

In trajectory-based motion planning, the objective is not only to identify a single feasible path but to characterize a distribution over feasible motion trajectories that satisfy system dynamics and task constraints. 
This requires a data generation mechanism that captures both environmental feasibility and goal-directed behavior. 
Rather than relying on hand-crafted motion primitives or deterministic planners, we construct this motion trajectory distribution empirically by learning a policy through RL.

We model the single-agent navigation problem as a Markov Decision Process (MDP), where the agent state is denoted by $x_t \in \mathbb{R}^d$ and the policy $\pi_\theta(a_t \mid x_t)$ induces motion trajectories $\tau = (x_0, x_1, \dots, x_H)$ through interaction with the environment. The policy is learned by maximizing the expected cumulative reward defined in Eq.~\eqref{eq:rl_objective}:
\begin{equation}
J(\theta) = \mathbb{E}_{\pi_\theta} \left[ \sum_{t=0}^{H} \gamma^t r(x_t, a_t) \right],
\label{eq:rl_objective}
\end{equation}
where $r(x_t, a_t)$ is the reward signal at state $x_t$ under action $a_t$, and $\gamma \in (0, 1]$ is a discount factor that controls the relative weight of future rewards.

This objective's role in our framework is not to produce a single optimal trajectory, but to induce a distribution over feasible motion trajectories. 
In particular, the learned policy induces a stochastic process over motion trajectories in which higher-reward trajectories are more likely to be sampled. As a result, the collected dataset is naturally biased toward feasible and goal-directed motion patterns, providing a meaningful basis for learning a motion trajectory distribution.

To capture long-term feasibility and goal-reaching behavior, we define the value function in Eq.~\eqref{eq:value_function}:

\begin{equation}
V^\pi(x_t) = \mathbb{E}_{\pi} \left[ \sum_{k=0}^{\infty} \gamma^k r(x_{t+k}, a_{t+k}) \mid x_t \right],
\label{eq:value_function}
\end{equation}
which satisfies the Bellman equation in Eq.~\eqref{eq:bellman}:

\begin{equation}
V^\pi(x_t) = \mathbb{E}_{\pi} \left[ r(x_t, a_t) + \gamma V^\pi(x_{t+1}) \right].
\label{eq:bellman}
\end{equation}
In this context, Eq.~\eqref{eq:value_function} implicitly encodes whether a state lies on a feasible motion trajectory toward the goal, thereby shaping the motion trajectory distribution toward valid navigation behaviors.

We train the policy using PPO \cite{schulman2017proximal}, which updates the policy by maximizing a clipped surrogate objective defined in Eq.~\eqref{eq:ppo_objective}:

\begin{equation}
\mathcal{L}_{\text{traj-policy}}
(\theta) =
\mathbb{E}_t \left[
\min \left(
r_t(\theta)\hat{A}_t,\,
\text{clip}(r_t(\theta), 1-\varepsilon, 1+\varepsilon)\hat{A}_t
\right)
\right],
\label{eq:ppo_objective}
\end{equation}
where $\hat{A}_t$ is the estimated advantage at timestep $t$, representing how much better action $a_t$ is relative to the average behavior of the policy, and $\varepsilon$ is a clipping hyperparameter that limits the size of policy updates. The probability ratio ($r_t(\theta)$) is defined in Eq.~\eqref{eq:ppo_ratio}:
\begin{equation}
r_t(\theta) = \frac{\pi_\theta(a_t \mid x_t)}{\pi_{\theta_{\text{old}}}(a_t \mid x_t)}.
\label{eq:ppo_ratio}
\end{equation}
where $\pi_{\theta_{\text{old}}}$ denotes the policy from the previous update step.
This objective stabilizes policy updates while preserving sufficient stochasticity in action selection, which enables the policy to generate diverse motions that explore multiple feasible motion patterns. 
In this study, such diversity is critical for constructing a dataset that provides broad coverage of the motion space, which in turn supports effective learning of a generative trajectory model.

After training, the learned policy is used as a motion trajectory generator. By executing the policy from different start-goal conditions, we collect a dataset as defined in Eq.~\eqref{eq:trajectory_dataset}:

\begin{equation}
\mathcal{D} = \{ \tau^{(k)} \}_{k=1}^{K}, \quad
\tau = (x_0, x_1, \dots, x_H),
\label{eq:trajectory_dataset}
\end{equation}
which provides an empirical approximation of the feasible motion distribution, capturing diverse and goal-directed motion patterns under the system dynamics. 
In the following subsection, we learn a generative model over motion trajectories based on this dataset.

\subsection{Single-Agent Motion Modeling via Diffusion}
Given the trajectory dataset $\mathcal{D}$ defined in Eq.~\eqref{eq:trajectory_dataset}, we aim to learn a probabilistic model $p_\theta(\tau)$ that captures the distribution of feasible motion trajectories. Unlike deterministic planners that generate a single solution, our objective is to learn a generative model that can represent diverse and multimodal motion patterns observed in the dataset.

To this end, we employ a conditional diffusion model, which learns to generate motion trajectories through an iterative denoising process. The model defines a forward diffusion process that gradually perturbs a motion trajectory $\tau_0$ by injecting Gaussian noise over $T$ steps. 
The forward transition is defined in Eq.~\eqref{eq:forward_process}:
\begin{equation}
q(\tau_t \mid \tau_{t-1}) = \mathcal{N}\left( \tau_t; \sqrt{1 - \beta_t}\, \tau_{t-1}, \beta_t I \right),
\label{eq:forward_process}
\end{equation}
where $\beta_t$ is a predefined variance schedule. 
This process results in a noisy motion trajectory $\tau_t$ that progressively loses its structure.
The marginal distribution of $\tau_t$ conditioned on the original trajectory $\tau_0$ can be expressed in closed form in Eq.~\eqref{eq:forward_marginal}:
\begin{equation}
q(\tau_t \mid \tau_0) = \mathcal{N}\left( \tau_t; \sqrt{\bar{\alpha}_t}\, \tau_0, (1 - \bar{\alpha}_t) I \right),
\label{eq:forward_marginal}
\end{equation}
where $\alpha_t = 1 - \beta_t$ and $\bar{\alpha}_t = \prod_{s=1}^{t} \alpha_s$. This formulation enables direct sampling of noisy trajectories during training.

The reverse process is parameterized by a neural network that learns to recover clean motion trajectories from noisy inputs. 
Instead of directly predicting the denoised trajectory, we adopt a noise prediction parameterization, where the model $\epsilon_\theta(\tau_t, t, c)$ estimates the noise added in the forward process, conditioned on $c = (x_{\text{start}}, x_{\text{goal}})$.
Under this parameterization, the reverse transition can be expressed as Eq.\eqref{eq:reverse_process}:
\begin{equation}
p_\theta(\tau_{t-1} \mid \tau_t, c) = \mathcal{N}\left( \tau_{t-1}; \mu_\theta(\tau_t, t, c), \Sigma_t \right),
\label{eq:reverse_process}
\end{equation}
where the mean is computed from the predicted noise as given in Eq.~\eqref{eq:reverse_mean}:
\begin{equation}
\mu_\theta(\tau_t, t, c) =
\frac{1}{\sqrt{\alpha_t}} \left(
\tau_t - \frac{1 - \alpha_t}{\sqrt{1 - \bar{\alpha}_t}} \epsilon_\theta(\tau_t, t, c)
\right).
\label{eq:reverse_mean}
\end{equation}

The diffusion-based motion generator is trained by minimizing the mean-squared error between the true noise and the predicted noise, as defined in Eq.~\eqref{eq:diffusion_loss}:
\begin{equation}
\mathcal{L}_{\text{traj-gen}}(\theta) =
\mathbb{E}_{\tau_0, \epsilon, t}
\left[
\| \epsilon - \epsilon_\theta(\tau_t, t, c) \|^2
\right],
\label{eq:diffusion_loss}
\end{equation}
where $\epsilon \sim \mathcal{N}(0, I)$ and $\tau_t$ is generated using Eq.~\eqref{eq:forward_marginal}.

This training objective enables the model to learn a trajectory distribution that captures the variability and structure of feasible motions observed in $\mathcal{D}$. 
During inference, motion trajectories are generated by iteratively applying the learned reverse process starting from Gaussian noise, while enforcing boundary conditions through conditioning on start and goal states.
As a result, the diffusion model serves as a generative prior over motion trajectories, producing diverse motion plans that remain consistent with the empirical distribution induced by the RL-generated dataset. 
This learned distribution $p_\theta(\tau)$ is subsequently extended to multi-agent settings in the next subsection.
%%%%%%%%%%%%%%%%%%%%%%%%%%%%%%%%%%%%%%%%%%%%%%%%%%%%%%%%%%%

\subsection{Decentralized Multi-Agent Trajectory Generation}

The single-agent diffusion model introduced in Section 4.2 provides a generative prior over feasible motion trajectories. 
We extend this model to a multi-agent setting by leveraging its ability to independently generate motion trajectories conditioned on agent-specific constraints.
Let $N$ denote the number of agents. For each agent $i$, we define its trajectory as $\tau_i = (x^i_0, x^i_1, \dots, x^i_H)$ and its conditioning variables as $c_i = (x^i_{\text{start}}, x^i_{\text{goal}})$. 
Using the shared diffusion model, each agent independently samples a motion trajectory according to Eq.~\eqref{eq:independent_sampling}:
\begin{equation}
\tau_i \sim p_\theta(\tau_i \mid c_i), \quad i = 1, \dots, N,
\label{eq:independent_sampling}
\end{equation}
where $p_\theta(\tau \mid c)$ is the motion trajectory distribution learned in Section 4.2.

Under this formulation, the joint distribution over multi-agent motion trajectories $X = (\tau_1, \tau_2, \dots, \tau_N)$ can be expressed in Eq.~\eqref{eq:joint_factorization}:
\begin{equation}
p_\theta(X) = \prod_{i=1}^{N} p_\theta(\tau_i \mid c_i),
\label{eq:joint_factorization}
\end{equation}
which implies that motion trajectory generation is fully factorized across agents. This factorization enables decentralized motion generation, as each agent samples its motion trajectory independently without requiring joint optimization or inter-agent communication.

From a generative modeling perspective, each agent follows an independent stochastic denoising process driven by the shared diffusion model. 
As a result, all agents generate motion trajectories that are consistent with the same learned motion distribution while remaining conditionally independent given their individual start-goal constraints.
This decentralized formulation preserves scalability, as the computational complexity grows linearly with the number of agents rather than exponentially with the joint state space. 
Moreover, sharing a common generative model ensures that all agents produce motion trajectories that adhere to the same feasibility constraints captured in the dataset $\mathcal{D}$.

While the joint distribution in Eq.~\eqref{eq:joint_factorization} captures feasible motion generation at the individual agent level, it does not explicitly account for interactions among agents in shared environments. 
To incorporate interaction-aware behavior, we introduce a coordination mechanism in the following section.

%%%%%%%%%%%%%%%%%%%%%%%%%%%%%%%%%%%%%%%%%%%%%%%%%%%%%%%%%%%
\subsection{Centralized Value Function for Coordination}
\label{sec:value_function}

The decentralized trajectory generation described in Section 4.3 produces feasible motion trajectories for individual agents but does not explicitly account for interactions among agents. 
To incorporate interaction-aware behavior, we introduce a centralized value function that evaluates the coordination quality of joint motion trajectories.

Let $X = (\tau^1, \tau^2, \ldots, \tau^N)$ denote the joint trajectory set of all agents. 
We define a centralized value function $V_\phi(X)$ that estimates the expected cumulative reward of the joint motion, as defined in Eq.~\eqref{eq:central_value}:
\begin{equation}
V_\phi(X) \approx \mathbb{E} \left[ \sum_{t=0}^{H} \gamma^t r^{\text{joint}}(X_t) \right],
\label{eq:central_value}
\end{equation}
where $r^{\text{joint}}(X_t)$ is a joint reward that captures coordination objectives such as collision avoidance and efficient navigation.

The value function is learned using MARL under a centralized training paradigm. 
Specifically, the critic is trained to minimize the mean-squared error defined in Eq.~\eqref{eq:value_loss}:
\begin{equation}
\mathcal{L}_{\text{coord-value}}(\phi) =
\mathbb{E}_X \left[
\left( V_\phi(X) - \hat{R}(X) \right)^2
\right],
\label{eq:value_loss}
\end{equation}
where $\hat{R}(X)$ denotes the empirical return obtained from multi-agent rollouts.

Unlike decentralized policies that rely solely on local observations, the centralized value function operates on the joint state $s = [o^1 \| o^2 \| \cdots \| o^N] \in \mathbb{R}^{N \cdot d_o}$ and captures interaction effects among agents. As a result, $V_\phi(X)$ provides a global coordination signal that reflects the collective behavior of the system.
In our framework, the value function does not directly generate actions. Instead, it serves as a scoring function over joint trajectories, which is used to guide the generative process described in the next subsection.
In this way, $V_\phi(X)$ defines a coordination-aware scoring function over joint motion trajectories, which is used to reshape the generative distribution in the subsequent guidance step.
%%%%%%%%%%%%%%%%%%%%%%%%%%%%%%%%%%%%%%%%%%%%%%%%%%%%%%%%%%%
\subsection{Value-Guided Multi-Agent Planning}
\label{sec:guidance}

The centralized value function introduced in Section~\ref{sec:value_function} provides a global measure of coordination quality over joint motion trajectories. 
We incorporate this signal into the diffusion sampling process to guide motion trajectory generation toward coordinated multi-agent behaviors.

Let $p_\theta(X)$ denote the joint motion trajectory distribution induced by the decentralized diffusion model in Eq.~\eqref{eq:joint_factorization}. 
To incorporate coordination, we define a guided distribution as given in Eq.~\eqref{eq:compositional}:
\begin{equation}
\tilde{p}_\theta(X)
\propto
p_\theta(X)\exp\!\left(\lambda\, V_\phi(X)\right),
\label{eq:compositional}
\end{equation}
where $\lambda$ is a guidance scale controlling the influence of the coordination signal. 
This formulation biases the sampling process toward motion trajectories with higher coordination value while preserving the feasibility encoded in the diffusion prior.

Taking the logarithm of Eq.~\eqref{eq:compositional} and computing its gradient with respect to $X$ results in Eq.~\eqref{eq:log_gradient}:
\begin{equation}
\nabla_X \log \tilde{p}_\theta(X)
=
\nabla_X \log p_\theta(X)
+
\lambda \nabla_X V_\phi(X),
\label{eq:log_gradient}
\end{equation}
which shows that the guided distribution combines the diffusion prior with the gradient of the value function.
During diffusion sampling, the term $\nabla_X \log p_\theta(X)$ is implicitly captured by the learned denoising model, while $\nabla_X V_\phi(X)$ provides a coordination signal that steers the generation process toward interaction-aware motion patterns. 
This leads to a modified reverse transition for each agent, as defined in Eq.~\eqref{eq:guided_reverse}:
\begin{equation}
\tau^i_{t-1}
\sim
\mathcal{N}\!\left(
\mu^i_t + \lambda\,\Sigma_t\,\nabla^i,\;
\Sigma_t
\right),
\label{eq:guided_reverse}
\end{equation}
where $\mu^i_t = \mu_\theta(\tau^i_t, t, c^i)$ is the denoising mean of agent $i$ at diffusion step $t$, $\Sigma_t$ is the diffusion covariance, and $\nabla^i$ is the component of the joint gradient $\nabla_X V_\phi(X_t)$ corresponding to agent $i$.
This formulation can be interpreted as augmenting the diffusion dynamics with an additional drift term that promotes coordination among agents while preserving the feasibility of motion trajectories captured by the learned generative prior.
Because the value gradient is computed over the joint trajectory $X_t$, coordination information is propagated across all agents without requiring explicit communication during planning. 
Boundary conditions are enforced via conditioning at each denoising step, ensuring that start and goal constraints remain satisfied throughout the sampling process.
Overall, this formulation integrates generative modeling and RL by treating motion generation as probabilistic sampling and coordination as distribution shaping, resulting in a scalable framework for decentralized yet coordinated multi-agent motion planning.

The complete procedure is summarized in Algorithm~\ref{alg:guided_sampling}. 
Algorithm~\ref{alg:guided_sampling} implements the value-guided diffusion process described in Eq.~\eqref{eq:log_gradient} and Eq.~\eqref{eq:guided_reverse}, combining decentralized trajectory generation with centralized coordination signals.
%%%%%%%%%%%%%%%%%%%%%%%%%%%%%%%%%%%%%%%%%%%%%%%%%%%%%%%%%%%
\begin{algorithm}
\caption{Value-Guided Multi-Agent Motion Generation}
\label{alg:guided_sampling}
\begin{algorithmic}[1]

\Require Diffusion model $p_\theta$, value function $V_\phi$
\Statex \hspace{2.3em} Conditions $c^i = (x^i_{\text{start}}, x^i_{\text{goal}})$
\Statex \hspace{2.3em} Guidance scale $\lambda$
\Ensure Joint trajectory $X_0 = (\tau^1_0, \ldots, \tau^N_0)$

\For{$i = 1, \ldots, N$}
    \State $\tau^i_T \sim \mathcal{N}(0, I)$ \Comment{Initialize from noise}
    \State $\tau^i_T \leftarrow \text{Inpaint}(\tau^i_T, c^i)$ \Comment{Enforce boundary conditions}
\EndFor

\For{$t = T, \ldots, 1$}

    \For{$i = 1, \ldots, N$} \Comment{Decentralized denoising}
        \State $\mu^i_t \leftarrow \mu_\theta(\tau^i_t, t, c^i)$
    \EndFor

    \State $X_t \leftarrow (\tau^1_t, \ldots, \tau^N_t)$
    \State $\nabla_X V_\phi(X_t) \leftarrow [\nabla^1, \ldots, \nabla^N]$ \Comment{Centralized gradient}

    \For{$i = 1, \ldots, N$} \Comment{Guided reverse step}
        \State $\tau^i_{t-1} \sim \mathcal{N}\!\left(
                   \mu^i_t + \lambda\,\Sigma_t\,\nabla^i,\;
                   \Sigma_t
               \right)$
        \State $\tau^i_{t-1} \leftarrow \text{Inpaint}(\tau^i_{t-1}, c^i)$
    \EndFor

\EndFor

\State \Return $X_0$

\end{algorithmic}
\end{algorithm}

%%%%%%%%%%%%%%%%%%%%%%%%%%%%%%%%%%%%%%%%%%%%%%%%%%%%%%%%%%%
\section{Experiments}
\label{sec:experiments}

\subsection{Experimental Setup}
\begin{figure*}[tbhp]
    \centering
    \includegraphics[width=0.8\textwidth]{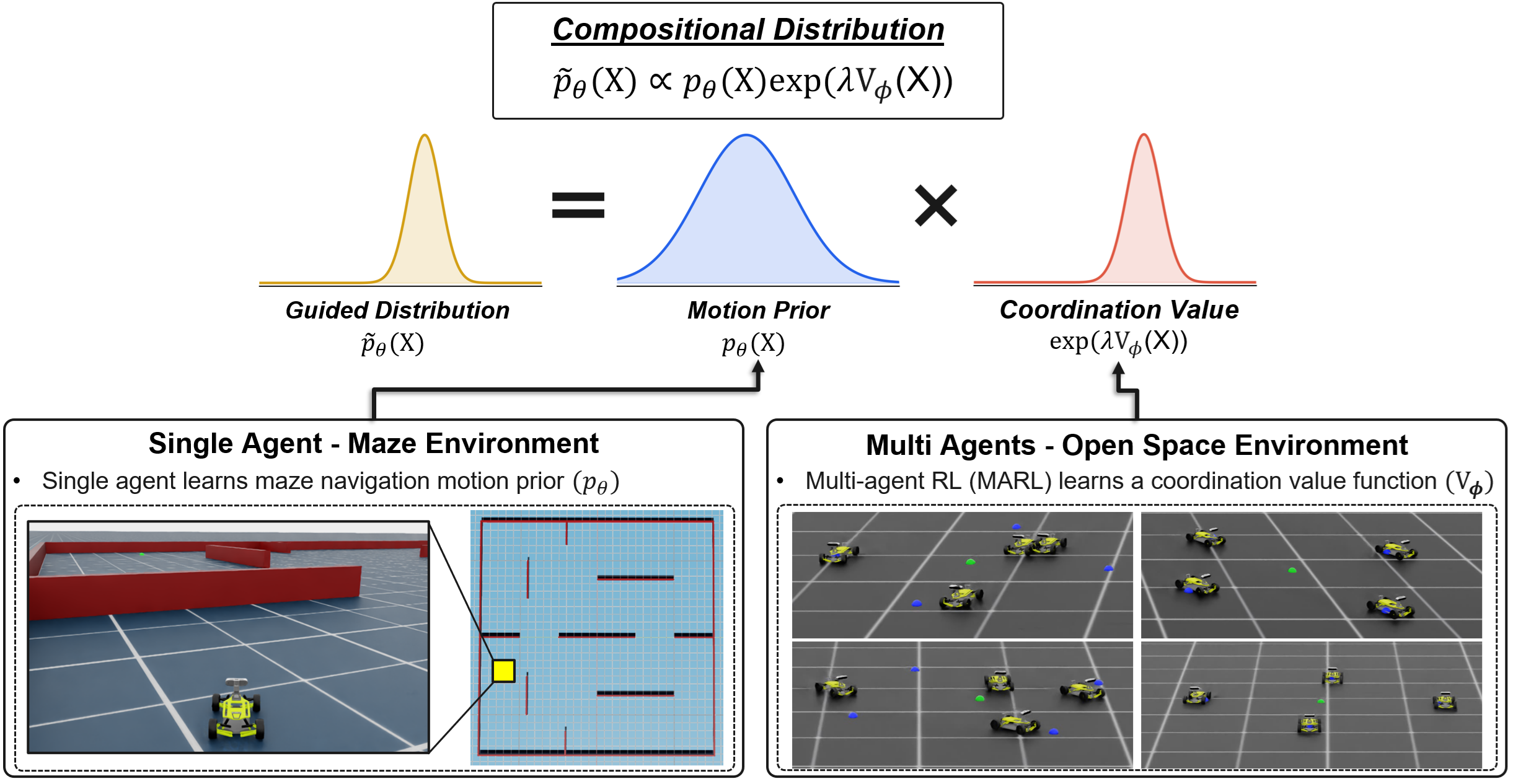}
    \caption{
        Experimental setup for the proposed framework, illustrated through the compositional distribution $\tilde{p}_\theta(X) \propto p_\theta(X)\exp(\lambda V_\phi(X))$.
        For the motion prior $p_\theta$, a single agent is trained via RL in a 30$\times$30\,m custom maze environment in Isaac Sim, where internal walls create constrained passages requiring feasible path planning between start and goal locations.
        The collected trajectories are used to train the diffusion model.
        For the coordination value $\exp(\lambda V_\phi(X))$, four agents are trained via multi-agent RL (MARL) in an open space environment, learning a centralized value function that estimates coordination quality from the joint observations of all agents.
        At inference time, the two components are composed without retraining, producing guided trajectories $\tilde{p}_\theta(X)$ in which multiple agents navigate the maze environment with reduced inter-agent interference.
    }
    \label{fig:experiment_setup}
\end{figure*}

All simulations were conducted in NVIDIA Isaac Sim 5.1.0 using the Isaac Lab framework. 
The reinforcement learning policies were implemented in PyTorch with CUDA support and trained using the skrl library on a system equipped with an NVIDIA RTX-4090 GPU.
The overall experimental setup consists of two separately trained components: a single-agent motion prior learned in a maze environment and a multi-agent coordination value learned in an open-space environment, as illustrated in Fig.~\ref{fig:experiment_setup}.

%%%%%%%%%%%%%%%%%%
%%%%%%%%%%%%%%%%%%

A custom maze spanning $30\,\mathrm{m} \times 30\,\mathrm{m}$, with coordinate ranges $[-15, 15]$ along both the $x$ and $y$ axes, was used for single-agent trajectory collection and guided multi-agent planning experiments. 
The maze contains internal walls that create constrained passages, requiring agents to plan feasible paths between distant start and goal locations. 
Each agent is modeled as a Leatherback robot, a four-wheeled vehicle with steering and a bounding radius of 
$0.254\,\mathrm{m}$.

\begin{figure}[thbp]
    \centering
    \includegraphics[width=\columnwidth]{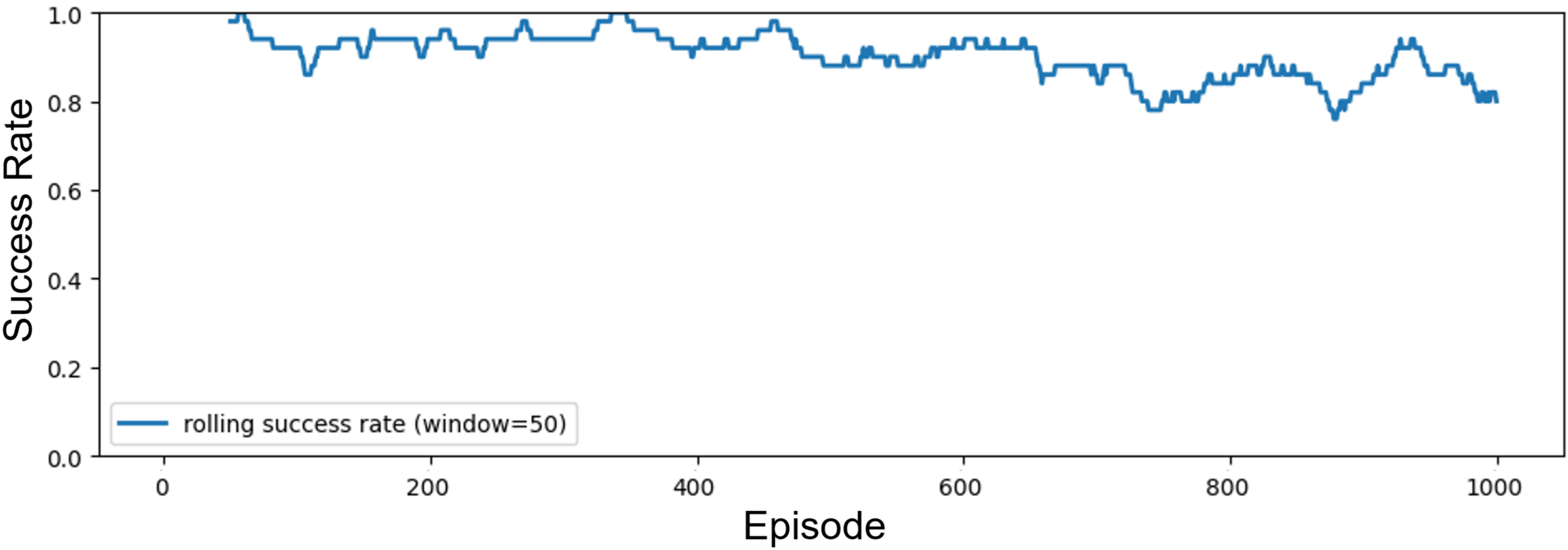}
    \caption{
        Rolling success rate (window = 50 episode) of the single-agent RL policy during online training. 
        %The success rate is computed over a moving window of recent episodes and remains relatively high throughout training.
    }
    \label{fig:single_rl_success}
\end{figure}

\begin{figure}[hbpt]
    \centering
    \includegraphics[width=0.75\columnwidth]{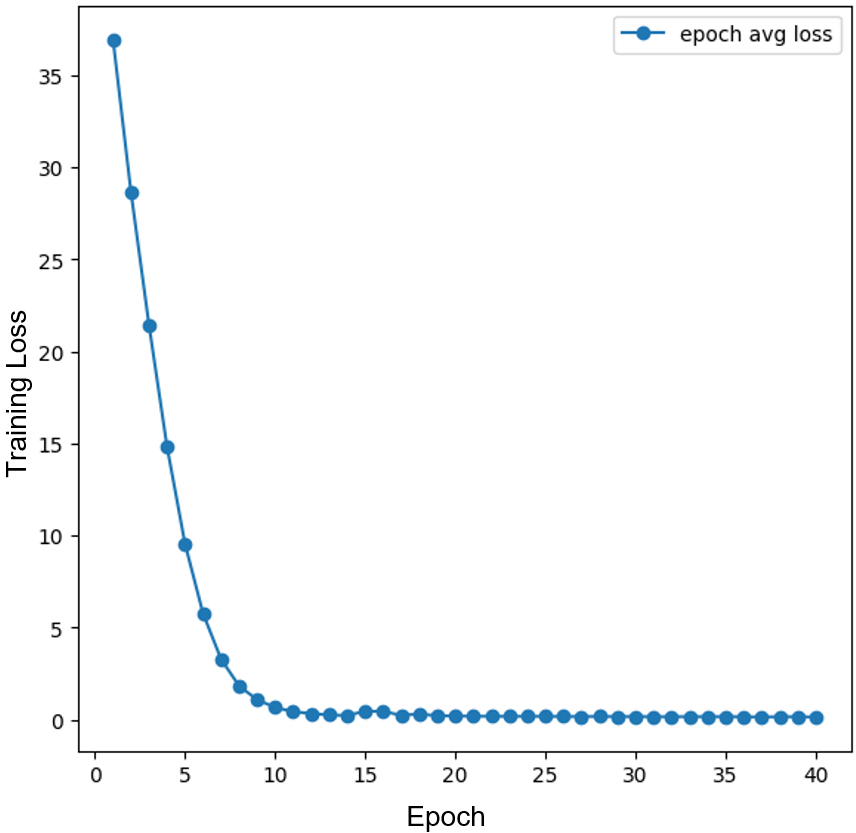}
    \caption{Training loss of the single-agent diffusion model over 40 epochs. 
    %The loss decreases rapidly in early epochs and stabilizes, indicating convergence of the trajectory generator.
    }
    \label{fig:diffusion_loss}
\end{figure}

To construct the motion prior, $p_\theta$, a single-robot navigation policy was first trained using PPO in the maze environment. 
The rolling success rate of the learned policy remains high during online training, suggesting that the policy provides sufficiently stable navigation behavior for trajectory collection, as shown in Fig.~\ref{fig:single_rl_success}.
The trained policy was then used to collect 2,800 navigation trajectories between start and goal locations. 
The diffusion planner was trained on this dataset with a planning horizon of 176 timesteps and 256 denoising steps, using a Temporal U-Net backbone to model the trajectory distribution.
The diffusion training loss decreases rapidly during the early stage of training and then stabilizes, suggesting stable training of the single-agent trajectory generator, as shown in Fig.~\ref{fig:diffusion_loss}.

\begin{figure}[hbpt]
    \centering
    \includegraphics[width=0.95\columnwidth]{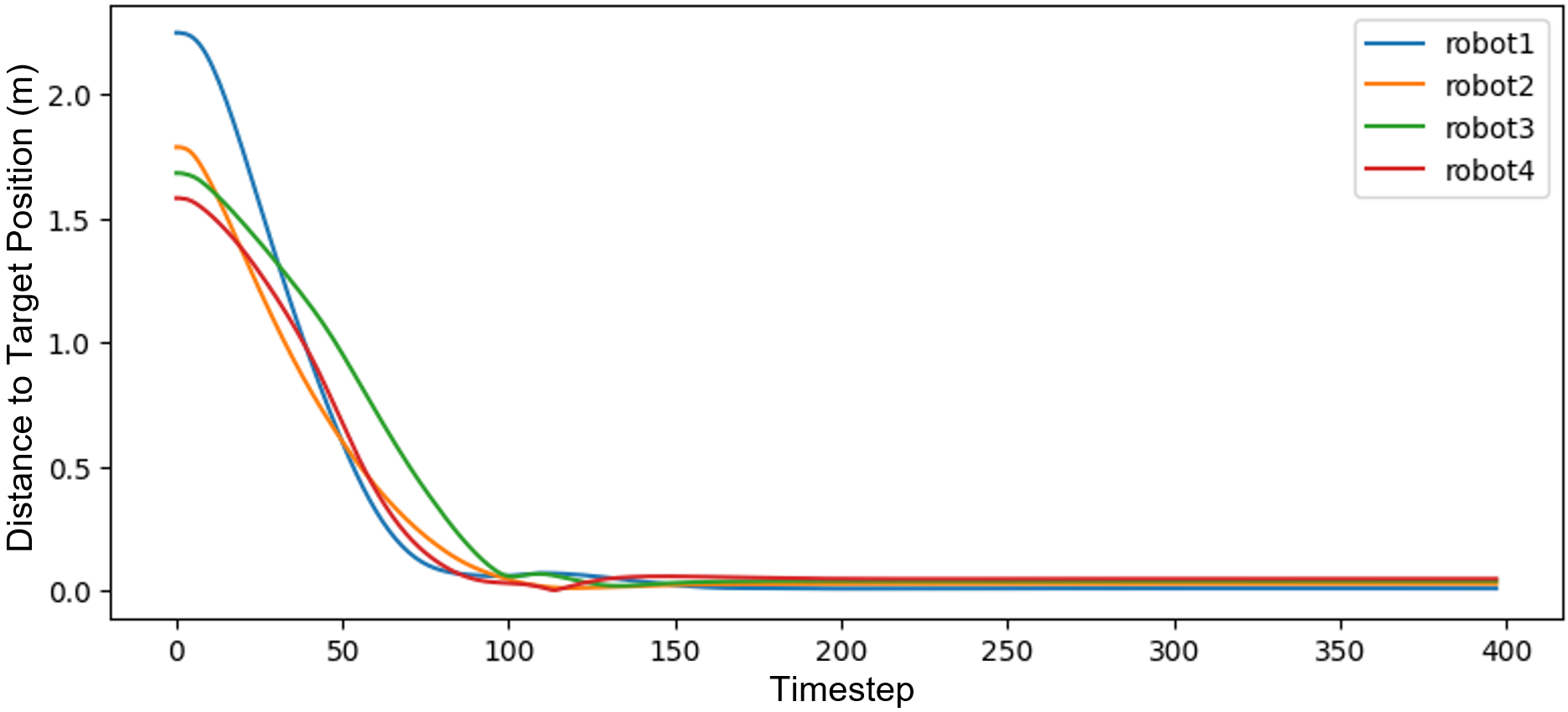}
    \caption{Distance from each robot to its assigned formation target in a representative MARL training example from the open-space environment. 
    %The distances decrease over time, indicating that the policy learns the intended relative-position coordination behavior.
    }
    \label{fig:marl_convergence}
\end{figure}

For multi-robot coordination, a MAPPO policy was separately trained in an open-space environment with four Leatherback robots navigating toward a shared target location.
Each robot was assigned a relative target position with respect to the group centroid, allowing the policy to learn coordinated motion while maintaining a desired spatial offset within the formation. 
This training setup provides a centralized value function $V_\phi$ that estimates coordination quality from the joint observations of all agents, but does not include maze walls or obstacle geometry.
The distance from each robot to its assigned formation target decreases over time, suggesting that the policy learns the intended relative-position coordination behavior, as shown in Fig.~\ref{fig:marl_convergence}.
At inference time, the centralized value function is used as a guidance signal to steer the diffusion sampling process in the maze, without retraining the diffusion model, producing coordinated multi-agent trajectories with reduced inter-agent interference.

\subsection{Evaluation Metrics}
\label{sec:metrics}
In multi-agent navigation, interactions among agents can affect the efficiency of task execution by introducing spatial constraints that lead to temporary path adjustments, motion slowdowns, or coordination delays. 
To capture these effects more broadly than direct collision events alone, this study evaluates navigation performance using interference events, defined as timesteps in which the distance between an agent and a nearby entity falls below a predefined safety threshold $\rho$.

In the maze environment considered in this work, interference may arise from both interactions between agents and interactions between an agent and environmental structures such as maze walls. 
Since the primary objective of this study is to evaluate inter-agent coordination, the main evaluation metric is defined based on agent-agent interference. 
Obstacle-related interference is additionally examined as an auxiliary metric to assess how the guided trajectories interact with environmental constraints. 
All interference rates are evaluated over the full planning horizon $H = 176$, corresponding to the trajectory length generated by the diffusion model.

The primary evaluation metric is the inter-agent interference rate, which measures how frequently agents experience proximity interactions with other agents during navigation, as in Equation~(\ref{eq:agent_interference}):
\begin{equation}
R_{agent} = \frac{T_{agent}}{T_{sim}} \times 100
\label{eq:agent_interference}
\end{equation}
, where $T_{agent}$ denotes the number of timesteps in which an agent is within the inter-agent interference threshold $\rho_{agent} = 0.508\,\text{m}$ of another agent, and $T_{sim}$ denotes the total number of timesteps in the planning horizon $H$.

As an auxiliary metric, the obstacle interference rate is additionally used to quantify the frequency of proximity interactions between agents and environmental structures such as maze walls, as in Equation~(\ref{eq:obstacle_interference}):
\begin{equation}
R_{obs} = \frac{T_{obs}}{T_{sim}} \times 100
\label{eq:obstacle_interference}
\end{equation}
, where $T_{obs}$ denotes the number of timesteps in which an agent is within the obstacle interference threshold $\rho_{obs} = 0.254\,\text{m}$ of an environmental obstacle.
%
%The total interference rate aggregates both types of interactions to provide an overall measure of spatial constraint during navigation, as in Equation~(\ref{eq:total_interference}):
%
%\begin{equation}
%R_{total} = \frac{T_{agent} + T_{obs}}{T_{sim}} \times 100
%\label{eq:total_interference}
%\end{equation}
%
%, where $T_{agent}$ and $T_{obs}$ correspond to the timestep counts defined above. 
Lower values of $R_{agent}$ indicate improved coordination and reduced inter-agent interference during navigation.
$R_{obs}$ is reported as a supplementary measure to assess how guidance affects trajectory proximity to environmental structures.

\subsection{Scenarios}
The proposed framework is evaluated in a multi-robot navigation task where four agents simultaneously traverse the maze toward assigned goal locations.
All robots are initialized near the lower-left region of the environment and are assigned goal positions near the upper-right region. 
To avoid identical trajectories and introduce spatial diversity, each robot is assigned an individual start and goal position offset from the group centroid by approximately 1.5\,m in each of the four orthogonal directions. 
This results in slightly shifted paths for each agent while preserving a common overall navigation direction.

To evaluate the effect of the proposed guidance mechanism, two planning approaches are compared. 
In the baseline setting, trajectories are generated solely by the diffusion planner without any guidance
signal. 
In the guided setting, the diffusion sampling process incorporates the centralized value function learned from multi-agent reinforcement learning, encouraging coordinated navigation among the agents.
Both approaches are evaluated using identical start-goal configurations, and the resulting trajectories are analyzed using the interference metrics described in Section~\ref{sec:metrics}.

%%%%%%%%%%%%%%%%%%%%%%%%%%%%%%%%%%%%%%%%%%%%%%%%%%%%%%%%%%%
\section{Results and Discussion}
\label{sec:results}

\subsection{Quantitative Interference Analysis}
\label{sec:quant}

Table~\ref{tab:interference_results} summarizes the inter-agent interference statistics defined in Section~\ref{sec:metrics}, comparing the unguided baseline diffusion planner with the proposed MARL-guided diffusion planner. 
For each condition, the experiment was repeated ten times, and the results were averaged.
The inter-agent interference rate $R_{agent}$ decreases from 55.4\% to 41.8\%, corresponding to a reduction of 13.6\% points.
%The obstacle interference rate $R_{obs}$ increases modestly from 4.6\% to 7.6\%.
%Despite this trade-off, the total interference rate $R_{total}$ decreases from 57.4\% to 47.9\%, indicating reduced spatial conflicts during multi-robot navigation.
These results suggest that value guidance improves coordination among independently generated diffusion trajectories.

%%%%%%%%%%%%%%%%%%%%%%%%%%%%%%%%%%
\begin{table}[h]
    \centering
    \caption{Mean inter-agent interference rates for baseline and MARL-guided diffusion.}
    \label{tab:interference_results}
    \begin{tabular}{lccc}
        \hline
        \textbf{Metric} & \textbf{Baseline [\%]} & \textbf{Guided [\%]} & \textbf{$\Delta$ [\%]} \\
        \hline
        $R_{agent}$ & 55.4 & 41.8 & $-$13.6 \\
        %$R_{obs}$   &  4.6 &  7.6 & $+$3.0  \\
        %\hline
        %\textbf{$R_{total}$} & \textbf{57.4} & \textbf{47.9} & \textbf{$-$9.5} \\
        \hline
    \end{tabular}
\end{table}
%%%%%%%%%%%%%%%%%%%%%%%%%%%%%%%%%%

\subsection{Qualitative Trajectory Comparison}

\begin{figure*}[thpb]
\begin{subfigure}[b]{\columnwidth}
\centering{
\includegraphics[width=0.84\linewidth]{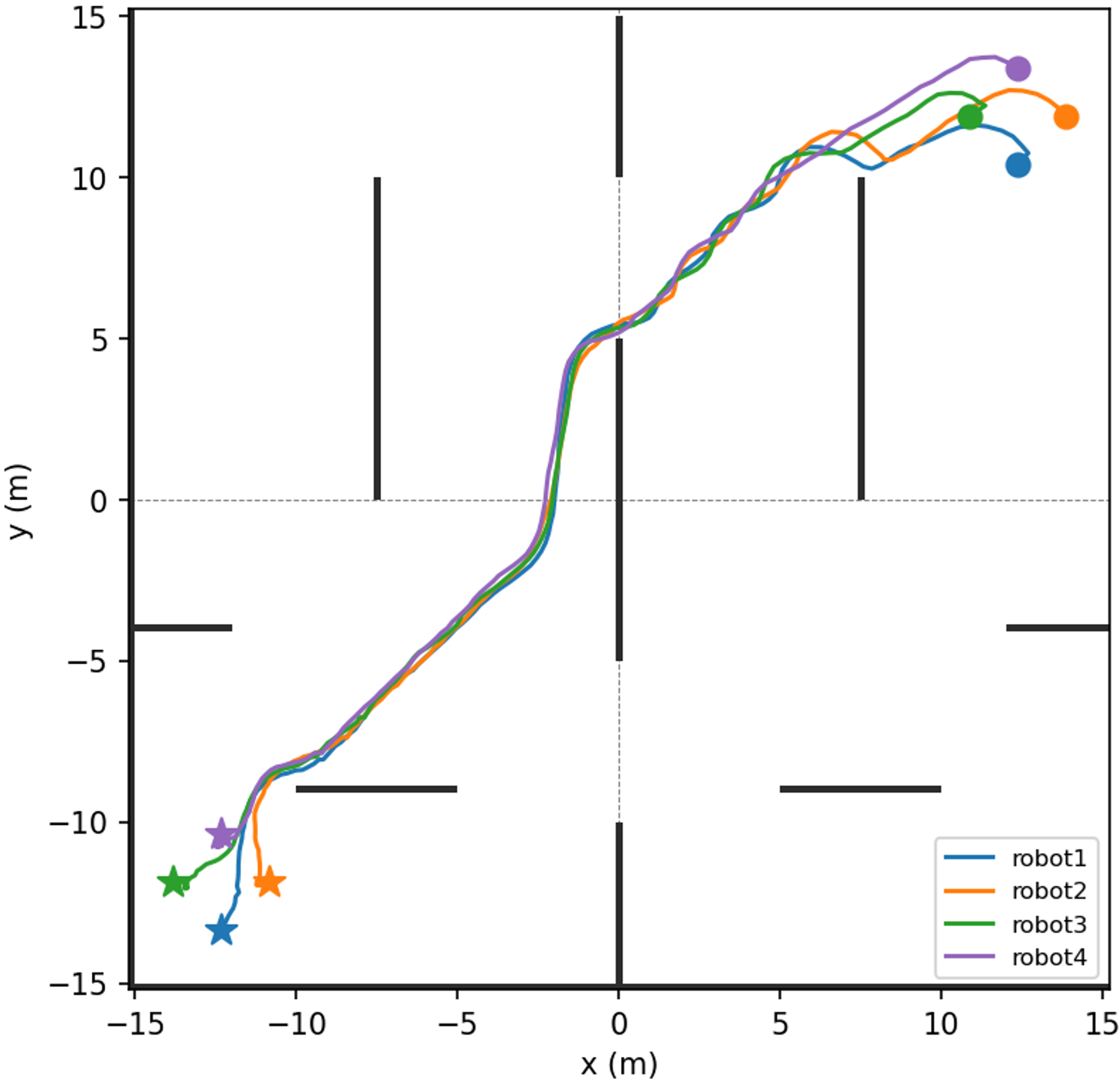}%
}
\subcaption{Baseline (Diffusion Only)\label{fig:traj_baseline}}
\end{subfigure}%
\hspace*{\columnsep}% with this space added, puts each figure at column center
%%%%%%%%%%%%% no line break between these two subfigures
\begin{subfigure}[b]{\columnwidth}
\centering{%
\includegraphics[width=0.85\linewidth]{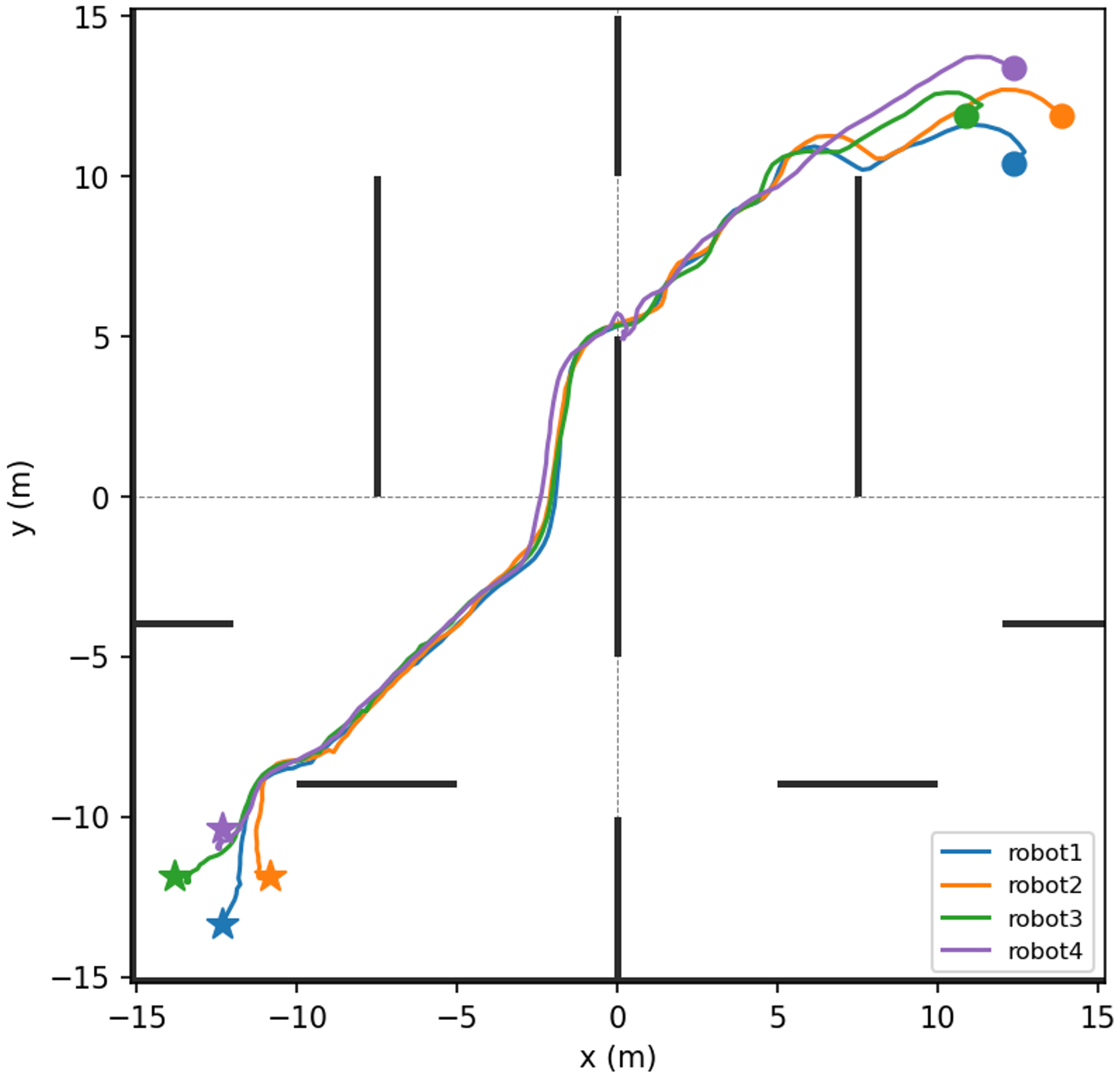}%
}%
\subcaption{MARL-Guided Diffusion\label{fig:traj_guided}}
\end{subfigure}
\caption{Comparison of multi-agent trajectories in the maze environment.
(a) Trajectories generated by the baseline planner using diffusion without guidance.
(b) Trajectories generated by the diffusion planner with MARL-based guidance.
Each color denotes one robot navigating from the lower-left start region to the upper-right goal region.
Star and circle markers indicate start and goal positions, respectively.
The guided planner produces greater spatial separation among agents while following a similar global route.
}
\label{fig:trajectories}
\end{figure*}

Figure~\ref{fig:trajectories} shows representative trajectories generated by the baseline diffusion planner and the MARL-guided planner. 
Each color corresponds to one agent navigating through the maze from the lower-left start region to the upper-right goal region.
In the baseline condition, agents travel through the corridor in a tightly clustered formation, with trajectories frequently overlapping along the same path. 
In contrast, the guided planner produces trajectories that follow a similar global route but maintain slightly
greater spatial separation among agents during traversal. 
This visual separation corresponds to a reduction in inter-agent interference, where $R_{agent}$ decreases from 59.7\% in the baseline case to 33.5\% with guidance.

Local trajectory adjustments can also be observed in the guided case.
For example, the trajectory of robot~4 (purple) briefly oscillates near the center of the maze corridor. 
Such deviations may reflect local corrections introduced by the guidance signal as it encourages greater spatial separation among agents during the reverse diffusion process. 
%The local corrections also cause guided trajectories to approach the maze boundaries more closely, which is consistent with the slightly higher obstacle interference rate observed in this case, $R_{obs}$ increasing from 2.3\% to 6.8\%, and with the overall quantitative results.

%%%%%%%%%%%%%%%%%  begin two column figure  %%%%%%%%%%%%%%%%%%%%%%%%%%%
%%%%%%%%%%%%%%%%%%%  end two column figure  %%%%%%%%%%%%%%%%%%%%%%%%%%

\subsection{Discussion}
\label{sec:discussion}

The reduction in $R_{agent}$ is consistent with the coordination objective encoded in the centralized value function.
The MARL critic is trained to maintain relative spacing among four agents navigating toward a shared goal.
During reverse diffusion, its value gradient provides a soft coordination signal that steers independently sampled trajectories away from regions of mutual spatial overlap.
This behavior is also visible in the qualitative trajectories shown in Fig.~\ref{fig:trajectories}, where agents that are tightly clustered under the baseline planner become more spatially separated under the guided planner while still following similar global routes through the maze.
Trajectories are generated by a shared single-agent diffusion model and conditioned on start-goal pairs offset from the group centroid.
As a result, guidance operates entirely at inference time without requiring joint trajectory modeling or explicit inter-agent communication.

As an auxiliary observation, we additionally examine the obstacle interference rate $R_{obs}$, which measures the frequency of proximity interactions between agents and maze walls with threshold $\rho_{obs} = 0.254\,\text{m}$.
In this analysis, $R_{obs}$ increased from 4.6\% to 7.6\%, suggesting a trade-off between reducing inter-agent interference and maintaining clearance from maze boundaries. 
The trade-off can be explained by the difference in training environments between the diffusion model and the MARL critic.
The diffusion planner is trained on trajectories collected in the maze environment and therefore implicitly captures obstacle avoidance behavior.
In contrast, the MARL critic is trained in an open-space multi-agent navigation setting and does not encode information about maze walls or obstacle geometry.
Consequently, value gradients that encourage stronger inter-agent separation may occasionally shift trajectories toward nearby boundaries, consistent with the wall proximity patterns observed in the guided trajectories in Fig.~\ref{fig:trajectories} (b).

The trajectory oscillations observed in the guided case reflect a related effect. 
The guidance signal introduces local corrections at each denoising step, and these iterative adjustments may produce path irregularities that do not appear in unguided sampling.
While such corrections contribute to reduced inter-agent interference, they represent a side effect of applying an open-space critic within a constrained maze environment.
Nevertheless, the substantial reduction in inter-agent interference indicates that value-based guidance can introduce effective multi-agent coordination during inference, even when trajectories are generated independently by a single-agent diffusion model.

%%%%% Conclusions %%%%%%%%%%%%%%%%%%%%%%%%%%%%%%%

\section{Concluding Remarks and Future Directions}
\label{sec:conclusion}
%(Conclusion)
This work presents a framework for coordinated multi-robot motion planning that combines a single-agent diffusion model with a centralized MARL value function for value-guided trajectory generation.
The proposed framework enables decentralized trajectory generation, allowing each robot to independently generate feasible and diverse candidate trajectories while introducing coordination through guidance.
By steering independently generated diffusion trajectories using value gradients from the MARL critic, the proposed approach reduces inter-agent interference without requiring joint trajectory modeling or retraining of the diffusion model.
Experimental results in a simulated maze environment demonstrate a reduction in the inter-agent interference rate from 55.4\% to 41.8\%, confirming that MARL-based value guidance can effectively coordinate independently generated trajectories at the planning stage.

%(Future Direction)
Several directions remain open for future investigation.
First, scaling the framework to larger agent populations would provide insight into the scalability of value-guided diffusion for multi-agent coordination.
Second, training or adapting the value function in obstacle-aware environments may help reduce the trade-off observed between improved inter-agent separation and increased proximity to maze boundaries. 
Third, validating the method on physical multi-robot systems represents an important step toward real-world deployment.
A particularly promising direction is the application of the proposed framework to multi-robot manufacturing environments.
In such settings, multiple robotic manipulators must coordinate their motions while satisfying manufacturability constraints, process dependencies, and shared workspace limitations.
Extending value-guided diffusion planning to these scenarios, including advanced manufacturing domains such as additive manufacturing and semiconductor fabrication, may enable coordinated multi-robot operation under complex process and geometric constraints.
Together, these directions highlight the potential of value-guided diffusion as a scalable framework for coordinated decision-making in decentralized multi-agent robotic systems.

%%%%% Acknowledgments %%%%%%%%%%%%%%%%%%%%%%%%%%%

\section*{Acknowledgments}
The authors would like to acknowledge the support provided by Arizona State University’s faculty startup fund and Fulton Fellowship.

%%%  REFERENCES  %%%%%%%%%%%%%%%%%%%%%%%%%%%%%%%%
%%
%% Put your references into your .bib file in the usual way. Run latex once, bibtex once, then latex twice.
%% The asmeconf.bst style allows @inproceedings and @proceedings to include: 
%%		venue = {Location of Conference}, 
%%		eventdate = {Month, days},

%\nocite{*}%% <=== Delete this line unless you want to typeset the entire contents of your .bib file !!

\bibliographystyle{asmeconf}  %% .bst file following ASME conference format. Do not change.
%\bibliography{bibliography}%% <=== change this to the name of your bib file

\end{document}